\documentclass[10pt,twocolumn,letterpaper]{article}

\usepackage{iccv}
\usepackage{times}
\usepackage{epsfig}
\usepackage{graphicx}
\usepackage{amsmath}
\usepackage{amssymb}
\usepackage{booktabs}

\makeatletter
\@namedef{ver@everyshi.sty}{}
\makeatother

\usepackage[pagebackref,breaklinks,colorlinks]{hyperref}

\usepackage[capitalize]{cleveref}
\crefname{section}{Sec.}{Secs.}
\Crefname{section}{Section}{Sections}
\Crefname{table}{Table}{Tables}
\crefname{table}{Tab.}{Tabs.}

\usepackage{xspace}
\usepackage{bm}
\usepackage{amsmath}
\usepackage{CJKutf8}
\usepackage{subcaption}
\usepackage{multirow}
\usepackage{bbding}
\usepackage[figuresright]{rotating}
\usepackage{makecell}
\usepackage{colortbl}

\usepackage{booktabs}

\usepackage{multirow}
\usepackage{ulem}
\usepackage{amsmath}
\usepackage{bbm}
\usepackage{stmaryrd}

\usepackage[dvipsnames]{xcolor}
\usepackage{enumitem}
\usepackage{microtype}
\usepackage{nicematrix}

\newcommand{\method}{TSGCNeXt\xspace}

\usepackage{newfloat}
\usepackage{listings}

\iccvfinalcopy 

\def\iccvPaperID{4888} 

\ificcvfinal\fi

\begin{document}

\title{TSGCNeXt: Dynamic-Static Multi-Graph Convolution for Efficient Skeleton-Based Action Recognition with Long-term Learning Potential}

\author{Dongjingdian Liu\\
{\tt\small DongjingdianLiu@cumt.edu.cn}
\and
Pengpeng Chen\textsuperscript{*}\\
{\tt\small chenp@cumt.edu.cn}
\and
Miao Yao\\
{\tt\small yaomiao@cumt.edu.cn}\\
\and
Yijing Lu\\ 
{\tt\small  luyijing@cumt.edu.cn}\\
\and
Zijie Cai\\
{\tt\small caizj@cumt.edu.cn}\\
\and
Yuxin Tian\\
{\tt\small yxtian@cumt.edu.cn}
}

\maketitle

\ificcvfinal\thispagestyle{empty}\fi

\begin{abstract}
Skeleton-based action recognition has achieved remarkable results in human action recognition with the development of graph convolutional networks (GCNs). 
However, the recent works tend to construct complex learning mechanisms with redundant training and exist a bottleneck for long time-series.
To solve these problems, we propose the Temporal-Spatio Graph ConvNeXt (TSGCNeXt) to explore efficient learning mechanism of long temporal skeleton sequences.
Firstly, a new graph learning mechanism with simple structure, Dynamic-Static Separate Multi-graph Convolution (DS-SMG) is proposed to aggregate features of multiple independent topological graphs and avoid the node information being ignored during dynamic convolution.
Next, we construct a graph convolution training acceleration mechanism to optimize the back-propagation computing of dynamic graph learning with 55.08\% speed-up.
Finally, the TSGCNeXt restructure the overall structure of GCN with three Spatio-temporal learning modules,efficiently modeling long temporal features. 
In comparison with existing previous methods on large-scale datasets NTU RGB+D 60 and 120, TSGCNeXt outperforms on single-stream networks. In addition, with the ema model introduced into the multi-stream fusion, TSGCNeXt achieves SOTA levels. On the cross-subject and cross-set of the NTU 120, accuracies reach 90.22\% and 91.74\%. The code is available at \url{https://github.com/vvhj/TSGCNeXt}.
\end{abstract}
\section{Introduction}
\label{sec:introduction}
Action recognition is a hot research topic of computer vision, widely used in human-computer interaction and security surveillance. Skeleton-based action recognition algorithms are gaining interest due to their robustness against background interference~\cite{weinland2011survey}. 

\begin{figure}
    \centering
     \begin{subfigure}[b]{0.235\textwidth}
         \centering
         \includegraphics[width=1.0\textwidth]{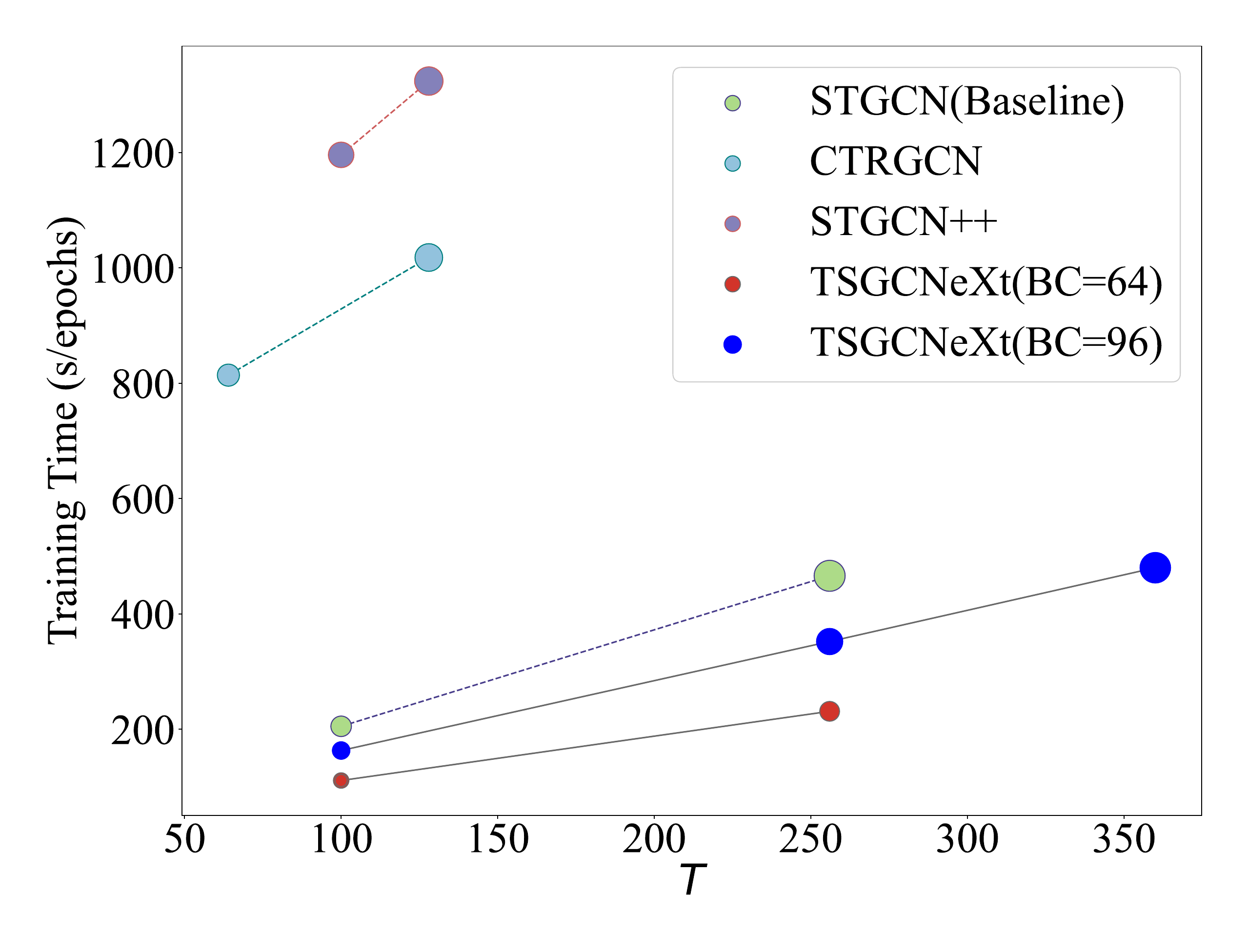}
         \caption{Training speed advantage}
         \label{1(a)}
    \end{subfigure}
    \begin{subfigure}[b]{0.235\textwidth}
         \centering
         \includegraphics[width=1.0\textwidth]{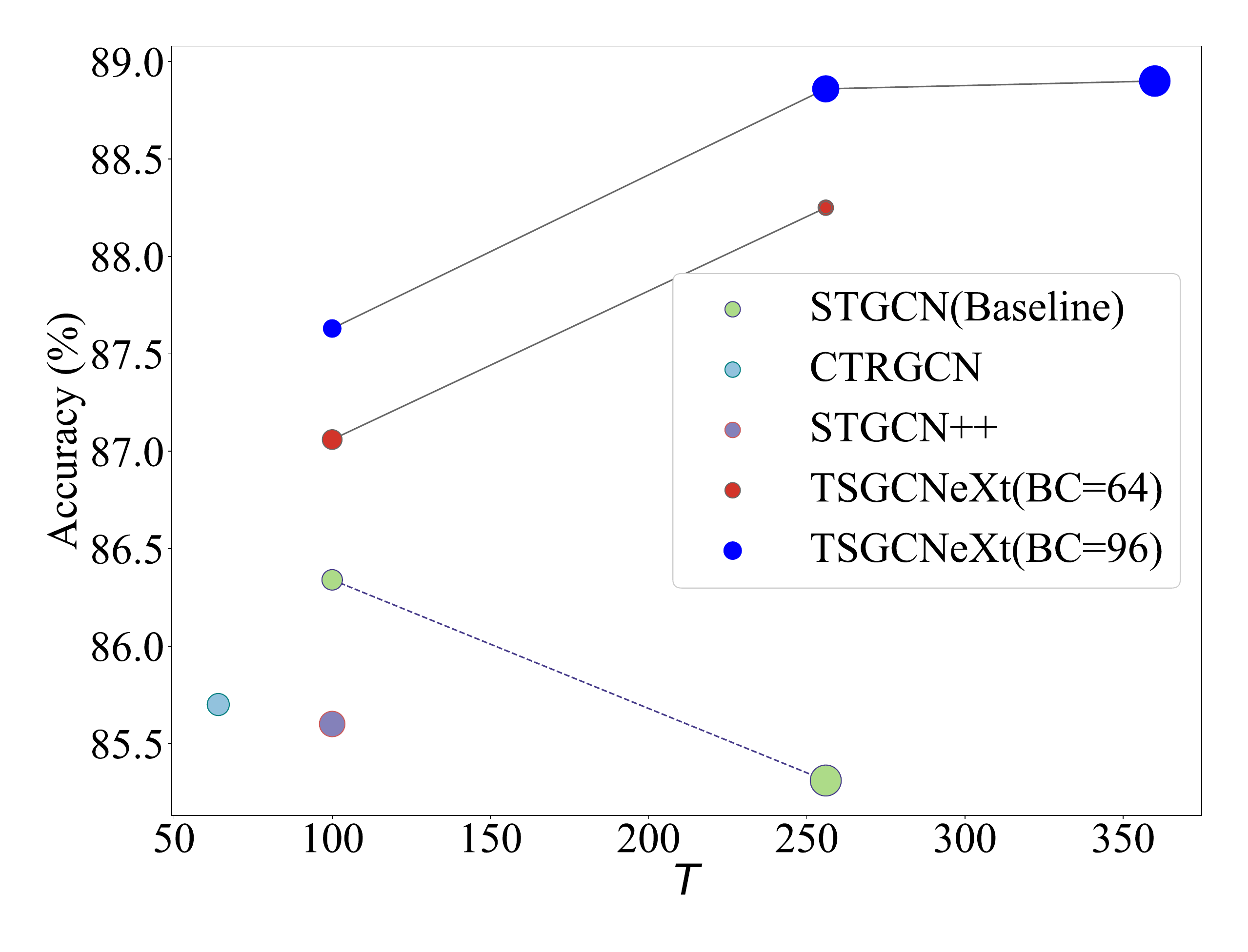}
         \caption{Accuracy advantage}
         \label{1(b)}
    \end{subfigure}
    \caption{Efficient training for long time-series. $T$ is the length of time-series. The size of the dot represents the number of occupied memory parameters. Unlike previous work that introduced complex learning mechanisms resulting in inefficient training and sacrificing long time series, TSGCNeXt can learn long time series information efficiently with fewer parameters. In addition TSGCNeXt can cope with the accuracy degradation caused by the increase of time-series data.}
    \label{fig:longtime}
    \vspace{-1em}
\end{figure}
The early skeleton-based action recognition methods utilize CNN~\cite{li2017skeleton} and RNN~\cite{du2015hierarchical} to model the spatio-temporal features of skeletal sequences, ignoring the topological relationships that exist between the joints~\cite{chen2021channel}. Since \cite{li2019actional} proposed STGCN to model skeletal data as a spatio-temporal graph structure, graph convolutional networks (GCNs) have become the dominant method. New approaches are essentially based on the STGCN framework. They focus on three main directions, the optimization of graph mechanisms, the fusion of multi-stream information and the introduction of new backbones. The optimization of the graph mechanism focuses on dynamic graphs~\cite{shi2019two,zhang2020semantics,obinata2021temporal,zeng2021learning}, multiple graphs~\cite{huang2020spatio,liu2020disentangling,ye2020dynamic} and non-shared topologies~\cite{cheng2020decoupling,chen2021channel}. These approaches often introduce new learning mechanisms, \eg self-attentive modules~\cite{ ye2020dynamic, chen2021channel}, making the network structure progressively more redundant~\cite{song2022constructing}. Multi-stream information fusion designs new representations of skeletal data, \eg bone, motion and angle. They use pre-fusion or post-voting mechanisms to enhance the recognition accuracy, but their single-stream representation capability is limited. The introduction of new backbone networks, \eg Shift-GCN~\cite{cheng2020skeleton}, EfficientGCN~\cite{song2022constructing}, Gest~\cite{bai2021gcst}, transforms the base modules and network structure of GCN at the combinatorial level by drawing on new backbone networks \cite{gudovskiy2017shiftcnn,tan2019efficientnet,vaswani2017attention}.

However, existing methods usually introduce more complex network structures in order to build new graph learning mechanisms, which brings greater training costs. Moreover, the length of time series is usually reduced to ensure the number of parameters and computational efficiency, losing the fine-grained temporal information. In addition, some methods reduce the accuracy when learning long time series, leaving a bottleneck in long time series learning. The recent architectural debate between CNN and Transformer~\cite{vaswani2017attention} has given rise to several new backbone networks, \eg Swin Transformer~\cite{liu2021swin} and ConvNeXt~\cite{liu2022convnet}. The continuous improvement of efficient computing structure makes the learning of long time series spatio-temporal data no longer constrained by computing efficiency.

To this end, we propose the Temporal-Spatio Graph ConvNeXt (TSGCNeXt) focusing on the graph learning mechanism, efficient GCN training and overall structure optimization for long time sequence learning. We first design a novel graph learning mechanism, the Dynamic-Static Separate Multi-graph Convolution (DS-SMG) module. After a separate convolution extracting temporal features, the DS-SMG takes pointwise convolution to aggregate the features (corresponding to SMG) of dynamic topology and self-static topology (corresponding to DS). SMG treats multiple adjacency matrices corresponding to the dynamic topology as independent learnable parameters, which avoids complex mechanisms to learn longer time series with fewer parameters. DS prevents some nodes from being overlooked after dynamic graph operations, improving spatial representation. Further, we propose a graph convolution training acceleration technique to improve the backward computing efficiency of graph representation without affecting learning. In addition, TSGCNeXt fuses the overall structure design of ConvNeXt with the Spatio-temporal graph convolution mechanism, introducing the larger base channel, the stage compute ratio adjustment and separable downsampling. Based on these, we design graph learning basic modules: TSGCNext block, Temporal Stem block and Separate Temporal Down Sampling (DS) block. These blocks can effectively aggregate the features of long time series, making TSGCNeXt suitable for long time series graph learning.

We perform experiments on multiple benchmarks: NTU60, NTU120 and NW-UCLA. Compared with previous works, results show that TSGCNeXt has a significant advantage in single-stream and training efficiency. With EMA models and different stage compute rate settings introduced into multi-stream fusion, the accuracy reaches the level of SOTA. On NTU120, Xsub and Xset reach 90.22\% and 91.74\%. On NTU60, X-sub accuracy is improved to 94.47\%. The main contributions of TSGCNeXt include:
\begin{itemize}
\item We design a novel graph learning mechanism named DS-SMG, effectively aggregating multiple topological information without complex attention mechanism.
\item A graph convolution training acceleration strategy is proposed to optimize the backward propagation efficiency, saving the learning cost.
\item With the reconstruction of architecture, TSGCNeXt efficiently aggregate long-term information, providing a paradigm for the long time series graph learning.
\item New multi-stream fusion approaches are proposed to take full advantage of the EMA model and different stage compute rate settings.
\end{itemize}
\section{Related Work}
\label{sec:related}
\subsection{GCN-based Skeleton Action Recognition}
\label{ssec:related_skeleton}
Yan \textit{et al.}\cite{yan2018spatial} first propose the ST-GCN introducing GCN into the skeleton-based action recognition. Subsequent efforts are proposed to investigate the following three directions:
\textbf{1) New graph representation.} \cite{li2019actional}, \cite{huang2020spatio} and \cite{liu2020disentangling} expand multi-adjacency of spatio-temporal graphs. \cite{shi2019two}, \cite{zhang2020semantics} and \cite{ye2020dynamic} learn dynamic-adjacency topology by introducing self-attention or context-enriched technology. \cite{obinata2021temporal} and \cite{zeng2021learning} extend the temporal graph to enable joints interaction across frames. \cite{cheng2020decoupling} and \cite{chen2021channel} study topology-non-shared mechanism.
\textbf{2) Multi-stream features.} \cite{song2020richly} decouple joints to multiple subsets and take multi-stream integration. \cite{li2019actional} introduce bone stream and train the two streams separately with softMax fraction summation. Further, \cite{shi2019skeleton} and \cite{qin2021fusing} introduce motion and angle stream. \cite{song2022constructing} early fuse multiple input streams with separable convolutional layers. Besides, \cite{duan2021revisiting} generate 2d skeletal data based on HRNet from video, bringing new paradigms.
\textbf{3) New backbone structures.} Works aim to draw on novel network architectures. \cite{cheng2020skeleton} proposes shift GCN to adjust receptive fields. \cite{song2022constructing} take Efficient as backbone to optimize parameters. \cite{bai2021gcst} introduce Transformer for skeleton action recognition. 

Our work simultaneously involves improvements in all areas, introducing the new backbone ConvNeXt, proposing a Dynamic-Static Separate Multi-graph Convolution mechanism, and exploiting multiple streams to improve accuracy.
\begin{figure*}[htp]
  \centering
  \includegraphics[width=0.9\textwidth]{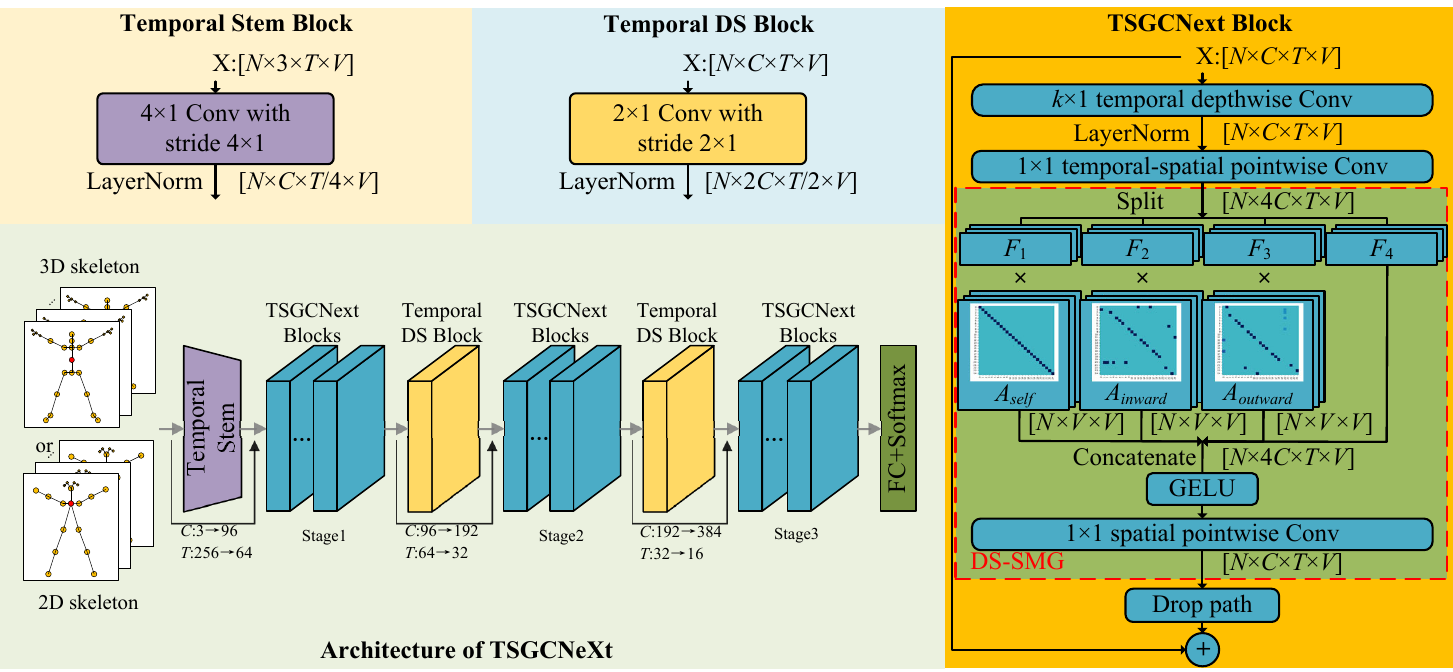}
  \caption{\label{fig:architecture}The overall pipeline of \method, including architecture and block details. The DS-SMG in red dashed box is the Dynamic-Static Separate Multi-graph Convolution. The compute ratio of three stage is designed to be $2:5:2$ or $4:3:2$.}
  \vspace{-1em}
\end{figure*}
\subsection{Modern Backbone Models}
\label{ssec:related_efficient}
Early backbone networks, VGG~\cite{simonyan2014very}, Inception~\cite{szegedy2015going}, ResNet~\cite{he2016deep}, and DenseNet~\cite{huang2017densely} study convolutional operator and computational processes. Nowadays, modern backbones mainly focus on following areas:
\textbf{1) convolutional architecture}. Some work focuses on the combinations of operators and hyperparameters of models. Efficientnet~\cite{tan2019efficientnet} and RegNet~\cite{radosavovic2020designing} search the depth, width, resolution and module combination using neural architecture search. Some research focuses on operator, such as the depth-wise separable convolution of MobileNet~\cite{howard2017mobilenets}, and convolutional kernel groupings in GhostNet~\cite{han2020ghostnet}.
\textbf{2) transformer mechanism.} ViT~\cite{dosovitskiy2020image} employs patch embedding to transform a visual feature adapting to the transformer. TNT~\cite{han2021transformer} create the Transformer-iN-Transformer for pixel-level modeling. These backbones have revolutionized CNN architecture, demonstrating the huge potential of global feature learning.
\textbf{3) transformer-convolution mechanism fusion.} Swin Transformer~\cite{liu2021swin} draws on the hierarchical design of CNNs to construct the network and designs a shift window attention. Following the Swin Transformer, ConvNeXt~\cite{liu2022convnet} reshapes convolutional blocks using large kernel depth separable convolutions without using the attention mechanism, opening up a new generation of CNN research. 

Associated with the GCN architecture, we select the design philosophy of ConvNeXt to build basic network and rethink the key factors making graph learning successful. 
\section{Preliminaries}
\label{sec:Preliminaries}
\subsection{Data Preprocessing}
\label{ssec:preprocessing}
The formats of data are normalized 3D skeleton pre-processing in \cite{chen2021channel} and 2D skeleton generated by HRNet\cite{duan2021revisiting}. We unify the skeleton sequence form as $\mathcal{X}\in {{\mathbb{R}}^{3\times M\times T\times V}}$, where $M$, $T$, $V$ denote the number of people, frames, and joints, respectively. Assume that $\mathcal{X}\left[ :,{{m}_{i}},{{t}_{i}},{{v}_{i}} \right]$ is the coordinate of the joint ${{v}_{i}}$ of the person ${{m}_{i}}$ in the frame ${{t}_{i}}$. For 3D skeleton, $\mathcal{X}\left[ :,{{m}_{i}},{{t}_{i}},{{v}_{i}} \right]\in \{x,y,z\}$ denotes the 3D coordinates. For 2D skeleton, $\mathcal{X}\left[ :,{{m}_{i}},{{t}_{i}},{{v}_{i}} \right]\in \{x,y,c\}$ denotes the 2D coordinates and the confidence of the joint. Then we format $\mathcal{X}$ to 4 input features: joint ${{\mathcal{X}}_{\text{j}}}$, bone ${{\mathcal{X}}_{\text{b}}}$, joint motion ${{\mathcal{X}}_{\text{jm}}}$ and bone motion ${{\mathcal{X}}_{\text{bm}}}$.

\subsection{Graph Convolution}
\label{ssec:graphconv}
Graph convolutional networks are widely used to model Non-Euclidean structural data. Generally, the skeleton data is organized as a graph $G=\left( \mathcal{V},E \right)$, where $\mathcal{V}$, $E$ denote the joints and bones respectively. In order to use the relationships between the joints to guide the learning of the model, the GCN usually relies on 2 key components: adjacency matrix $\mathbf{A}\in {{\mathbb{R}}^{V\times V\times k}}$ and learnable convolutional weight matrix (kernels) $\mathbf{W}\in {{\mathbb{R}}^{{{\mathcal{C}}_{i}}\times {{\mathcal{C}}_{o}}\times S}}$. $k$ corresponds to different partitions. $S$,${{\mathcal{C}}_{i}}$,${{\mathcal{C}}_{o}}$ are kernel size, input channels and output channels. A GCN layer is defined as follows:
\begin{equation}
  \mathbf{Z}=\sigma \left( \sum\limits_{s=1}^{S}{\mathbf{\hat{A}X}{{\mathbf{W}}_{s}}} \right),
  \label{quez}
\end{equation}
,where $\mathbf{\hat{A}}$ is the normalized adjacency matrix, $\sigma \left( \cdot  \right)$ is the activation function, $\mathbf{X}\in {{\mathbb{R}}^{k\times V\times T\times {{\mathcal{C}}_{i}}}}$ is the input features and $\mathbf{Z}\in {{\mathbb{R}}^{V\times T\times {{\mathcal{C}}_{o}}}}$ is the output features. $\mathbf{A}$ is used to construct node-to-node relationships step-by-step after power multiplication, while $\mathbf{W}$ is used for feature transformation. If $S>1$, $\mathbf{W}$ aggregate the feature of neighbouring nodes.

\section{Temporal-Spatio Graph ConvNeXt}
This section introduces details to the TSGCNeXt as shown in the Fig.~\ref{fig:architecture}.
Firstly, we propose a new graph representation mechanism, Dynamic-Static Separate Multi-graph Convolution (DS-SMG).
Then, a general graph convolution acceleration strategy is proposed to save training costs.
Finally, the reconstruction of the overall architecture with three spatio-temporal learning blocks are presented to realize efficient learning for long time series.

\begin{figure*}
    \centering
    \begin{subfigure}[b]{0.22\textwidth}
         \centering
         \includegraphics[width=1.0\textwidth]{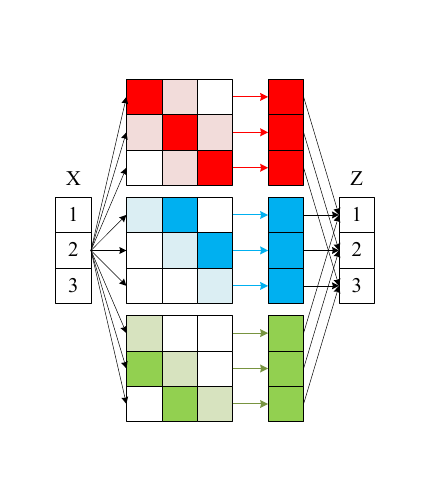}
         \caption{Classical GLM}
         \label{(a)}
    \end{subfigure}
    \begin{subfigure}[b]{0.22\textwidth}
         \centering
         \includegraphics[width=1.0\textwidth]{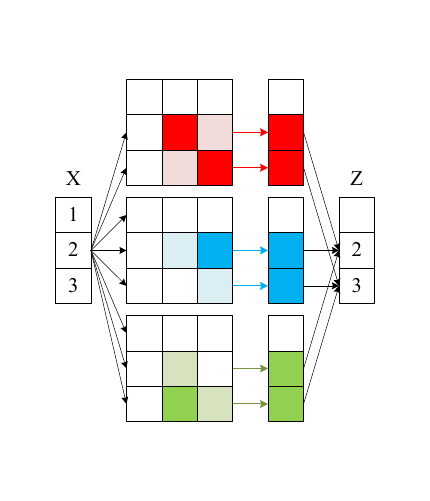}
         \caption{Node information lost}
         \label{(b)}
    \end{subfigure}
    \begin{subfigure}[b]{0.22\textwidth}
         \centering
         \includegraphics[width=1.0\textwidth]{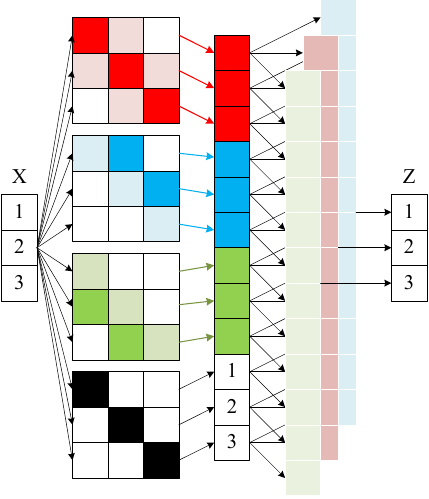}
         \caption{DS-SMG}
         \label{(c)}
    \end{subfigure}
    \begin{subfigure}[b]{0.22\textwidth}
         \centering
         \includegraphics[width=1.0\textwidth]{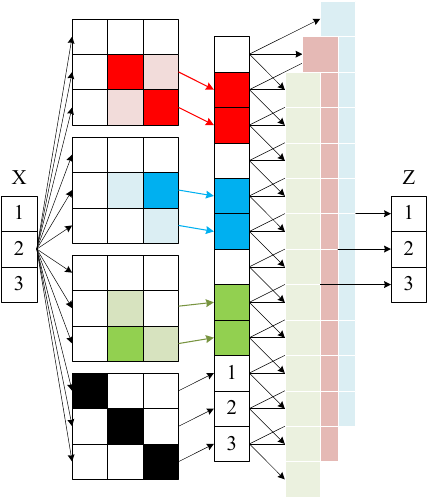}
         \caption{No node information lost}
         \label{(d)}
    \end{subfigure}
    \caption{Graph learning mechanism visualization. (a) and (b) correspond to the classical graph learn mechanism (GLM). (c) and (d) correspond to our DS-SMG mechanism. (b) and (d) are the special cases when the graph relationship is sparse.} 
    \label{fig:complexity}
    \vspace{-1em}
\end{figure*}
\subsection{DS-SMG}
\label{ssec:smc}

The Multi-graph adjacency matrix is one of the most important techniques used by skeletal graph convolution to learn relationships between nodes. Our $\mathbf{A}$ is denoted as:
\begin{equation}
  \mathbf{A}=\left\{ {{A}_{s}},{{A}_{i}},{{A}_{o}} \right\},
\end{equation}
where ${{A}_{s}}$, ${{A}_{i}}$, ${{A}_{o}}$ correspond to the self, inward, and outward relationships. Thus in the classical graph convolution architecture, Equ.\ref{quez} is expressed as:
\begin{equation}
  \mathbf{Z}=\sigma \left( {{{\hat{A}}}_{s}}\mathbf{XW}+{{{\hat{A}}}_{i}}\mathbf{XW}+{{{\hat{A}}}_{o}}\mathbf{XW} \right),
\end{equation}
which aggregate multi-adjacency matrixes with multicollinearity. It allows for a linear relationship between different topologies rather than topology-non-shared. In CTR-GCN, topology-non-shared can enhance graph learning. So we design the Dynamic-Static Separate Multi-graph Convolution to explore independent multi-adjacency matrix information aggregation. The mechanism is as follows:
\begin{equation}
    \mathbf{Z}=\left[ {{{\hat{A}}}_{s}}\sigma \left( F_1 \right),{{{\hat{A}}}_{i}}\sigma \left( F_2 \right), {{{\hat{A}}}_{o}}\sigma \left( F_3 \right),\sigma \left( F_4 \right) \right] \mathcal{W}\gamma,
\end{equation}
where ${{\mathbf{W}}^{j}}\in {{\mathbb{R}}^{{{\mathcal{C}}_{i}}\times {{\mathcal{C}}_{i}}}}$ corresponds to $\mathbf{W}[:,\left( j-1 \right){{\mathcal{C}}_{i}}:j{{\mathcal{C}}_{i}}]$, $F_i = \sigma \left( \mathbf{X}{{\mathbf{W}}^{i}} \right)$, and $\mathcal{W}\in {{\mathbb{R}}^{4{{\mathcal{C}}_{i}}\times {{\mathcal{C}}_{i}}}}$ corresponds to spatial pointwise conv. The input $\mathbf{X}$ is first mapped by a multi-adjacency matrix to generate separate features for each adjacency. These independent features are then aggregated by spatial pointwise conv to generate a new point set $\mathbf{Z}$. It is worth noting that the features are divided equally into four copies. The first three copies are used to learn the topological relationships of individual nodes under different initialised topologies, while the last copy $F4$ has no matrix.

To visualize the advantages of the Separate Multi-graph, we plot a relational mapping of $\mathbf{Z}$ concerning $\mathbf{X}$. Here we consider $\mathbf{A}$ to be the learnable parameter. The Dynamic-Static Separate Multi-graph Convolution in Fig~\ref{(c)} contains features in each node from all topological perspectives compared to the classical graph convolution in Fig~\ref{(a)}, enabling more complex node-to-node relationships to be learned. In some cases, the information of some nodes in $\mathbf{X}$ cannot be transferred to $\mathbf{Z}$ due to the learnable adjacency matrix as shown in Fig~\ref{(b)}. In contrast, in Fig~\ref{(d)}, $F_4$ ensures that $\mathbf{Z}$ always contains information about all nodes of $\mathbf{X}$.

\subsection{Graph Convolution Training Acceleration}
Compared with CNNs, the most prominent feature of GCNs is the matrix multiplication between adjacent matrix and feature. This process can be expressed by einsum (Einstein summation convention). For adjacency matrix and features, the output Y of graph convolution operation with $k$ subgraphs can be expressed as:
\begin{equation}
    \mathbf{Y}=\text{enisum}\left( \text{''}VUk,NCTVk\to NCTUk\text{''},\mathbf{A},\mathbf{X} \right).
\end{equation}
In the experiment, we find that the training time is mainly concentrated in the backward calculation of the operation. For this reason, we design a backward computing friendly equivalent operation for training:
\begin{equation}
    \mathbf{\tilde{A}}=\text{repeat}\left( \mathbf{A} \right)=\left\{ {{{\mathbf{\tilde{A}}}}_{1}},{{{\mathbf{\tilde{A}}}}_{2}},\ldots ,{{{\mathbf{\tilde{A}}}}_{N}}\left| {{{\mathbf{\tilde{A}}}}_{i}}=\mathbf{A} \right. \right\},
\end{equation}
\begin{equation}
    \mathbf{\tilde{Y}}=\text{enisum}\left( \text{''}NVUk,NCTVk\to NCTUk\text{''},\mathbf{\tilde{A}},\mathbf{X} \right),
\end{equation}
where $\text{repeat}\left( \mathbf{A}\right)$ repeating $N$ times of $\mathbf{A}$. 

First, we prove the equivalence of $\mathbf{Y}$ and $\mathbf{\tilde{Y}}$. In the forward process:
\begin{equation}
\begin{split}
  \mathbf{\tilde{Y}}=\mathbf{\tilde{A}X}=\left[ {{{\mathbf{\tilde{A}}}}_{1}}{{\mathbf{X}}_{1}},{{{\mathbf{\tilde{A}}}}_{2}}{{\mathbf{X}}_{2}},\ldots ,{{{\mathbf{\tilde{A}}}}_{N}}{{\mathbf{X}}_{N}} \right] \\ 
 =\left[ \mathbf{A}{{\mathbf{X}}_{1}},\mathbf{A}{{\mathbf{X}}_{2}},\ldots ,\mathbf{A}{{\mathbf{X}}_{N}} \right]=\mathbf{AX}=\mathbf{Y}. 
\end{split}
\end{equation}
In backward, since $\mathbf{A}$ is a learnable parameter, we only need to show that the gradient of $\mathbf{A}$ in $\mathbf{\tilde{Y}}$ and $\mathbf{Y}$ is consistent:
\begin{equation}
\begin{split}
  \tfrac{\partial \mathbf{\tilde{Y}}}{\partial \mathbf{A}}=\tfrac{\partial \mathbf{\tilde{A}X}}{\partial \mathbf{A}}=\tfrac{1}{N}\sum\limits_{i=1}^{N}{\tfrac{\partial {{{\mathbf{\tilde{A}}}}_{i}}{{\mathbf{X}}_{i}}}{\partial {{\mathbf{A}}_{i}}}} \\ 
 =\tfrac{1}{N}\sum\limits_{i=1}^{N}{\tfrac{\partial {{\mathbf{A}}_{i}}{{\mathbf{X}}_{i}}}{\partial {{\mathbf{A}}_{i}}}}=\tfrac{\partial \mathbf{AX}}{\partial \mathbf{A}}=\tfrac{\partial \mathbf{Y}}{\partial \mathbf{A}}.
\end{split}
\label{equ:9}
\end{equation}
Thus, the parameter updates of $\mathbf{A}$ for $\mathbf{Y}$ and $\mathbf{\tilde{Y}}$ are consistent during training.

Next, we discuss that $\mathbf{\tilde{Y}}$ is more efficient than $\mathbf{Y}$ in backward process. Equ.~\ref{equ:9} shows that the backward calculation of $\mathbf{Y}$ is  a matrix-to-matrix derivative $\tfrac{\partial \mathbf{Y}}{\partial \mathbf{A}}$. However, for $\mathbf{\tilde{Y}}$, the actual gradient calculation follows the chain rule:
\begin{equation}
    \tfrac{\partial \mathbf{\tilde{Y}}}{\partial \mathbf{A}}=\tfrac{\partial \mathbf{\tilde{Y}}}{\partial \mathbf{\tilde{A}}}\tfrac{\partial \mathbf{\tilde{A}}}{\partial \mathbf{A}}.
\end{equation}
$\tfrac{\partial \mathbf{\tilde{A}}}{\partial \mathbf{A}}$  corresponds to $\text{repeat}\left( \mathbf{A}\right)$, the calculation time of which can be ignored. The time cost of $\mathbf{\tilde{Y}}$ is also a matrix-to-matrix derivative $\tfrac{\partial \mathbf{\tilde{Y}}}{\partial \mathbf{\tilde{A}}}$.  Athough $\mathbf{Y}$ and $\mathbf{\tilde{Y}}$ have the same dimensions,  $\mathbf{\tilde{A}}$ has a more dimension corresponding to the batch than $\mathbf{A}$.  In the parallel computation of matrix elements, in the backward computation of enisum, $\tfrac{\partial \mathbf{Y}}{\partial \mathbf{A}}$ needs $NC$ derivatives, while $\tfrac{\partial \mathbf{\tilde{Y}}}{\partial \mathbf{\tilde{A}}}$ only needs  $C$ times. Therefore, the calculation cost of $\mathbf{\tilde{Y}}$ is lower than that of $\mathbf{Y}$.  It is worth noting that $\mathbf{\tilde{Y}}$ is only used for training, because it will waste part of time on repeat operation in the forward process. 

\subsection{Reconstruction of the overall architecture}
\label{ssec:architecture}
The overall design of most of the previous GCNN-based networks for skeleton action classification graphs is based on STGCN. Blocks are divided into 3 stages and the number of base channels is 64. Inspired by ConvNeXt, the number of base channels is first set to 96 to increase the richness of learning features. Then we explore the effect of stage compute ratios(SCR) with a constant number of layers. The TSGCNeXt utilizes two main configurations: $2:5:2$ and $4:3:2$. $2:5:2$ is following the consensus that the heavy number of stage 3 is considered more effective for the visual backbones with 4 stages. $4:3:2$ accords to the principle of previous GCNNs. Both configurations are trained on different datasets to achieve better results.

Unlike previous work only contains spatio-temporal graph convolution blocks and classifications, TSGCNeXt introduces the principles of Patchify Stem, Separate Down Sampling, and CovNeXtify block of ConvNeXt~\cite{liu2022convnet}, designing Temporal Stem block, Separate Temporal Down Sampling block, and TSGCNeXt block respectively.

\noindent\textbf{Temporal Stem block.} The stem layer is used to transform input features to appropriate features by downsampling. However, in a typical skeleton GCN, the temporal features are usually extracted directly without using Stem, limiting the granularity of temporal information. Therefore, the Temporal Stem block expand the temporal dimension of the input to learn long-term temporal features. As shown in Fig~\ref{fig:architecture}, Temporal Stem block consists of $4\times 1$ Conv layer with stride $4\times 1$ and LayerNorm layer. Sequence length $T$ is set to 256 to ensure that the temporal feature dimensions correspond to comparable computational complexity with previous works. Since the convolution in the Temporal Stem block is non-overlapping, the "patchify" design is more efficient in constructing features.

\noindent\textbf{Separate Temporal Down Sampling block.} Downsampling is tightly coupled with spatio-temporal convolution layers in previous GCNs at the final layer of the stage. However, this often conflicts with the residual structure, as the downsampled features mismatch the input features. Thus, a Separate Down Sampling block is used in modern backbone structures by adding downsampled convolution module between stage and stage. As shown in Fig~\ref{fig:architecture}, the kernel size and stride of the Separate Temporal Down Sampling block are all $2\times 1$, aggregating features in the temporal dimension.

\noindent\textbf{TSGCNeXt block.} The module design template for the TSGCNeXt block follows the ConvNextfiy block with a large kernel depthwise convolution and MLPfiy pointwise convolution, as shown in Fig~\ref{fig:architecture}. In addition, the activation function and normalization layers use GLUE and LN to complement the distributed training strategy. Unlike the previous direct replacement of the new module with separate spatial and temporal convolution modules, we organically integrate the process of spatio-temporal information extraction into a single ConvNextfiy module. Previous spatio-temporal graph convolution modules tend to extract spatial features first and then temporal features. However, we extract the temporal features before extracting the spatial features, hence the term temporal-spatio (TS) convolution.
Specifically, the first layer is a $k\times 1$ temporal deepthwise convolution, where $k$ corresponds to the temporal receptive field. Each convolutional kernel is independently connected to a set of input and output channels. It reduces the number of parameters and complies with the topology-non-shared mechanism. Then $1\times 1$ temporal-spatial pointwise convolution establishes relationships between channels while expanding the dimensionality of features. We next take the Dynamic-Static Separate Multi-graph Convolution mechanism to extract spatial features and add drop paths to enhance the nonlinear modeling capability of the module.
\begin{figure}
    \centering
    \includegraphics[width=0.4\textwidth]{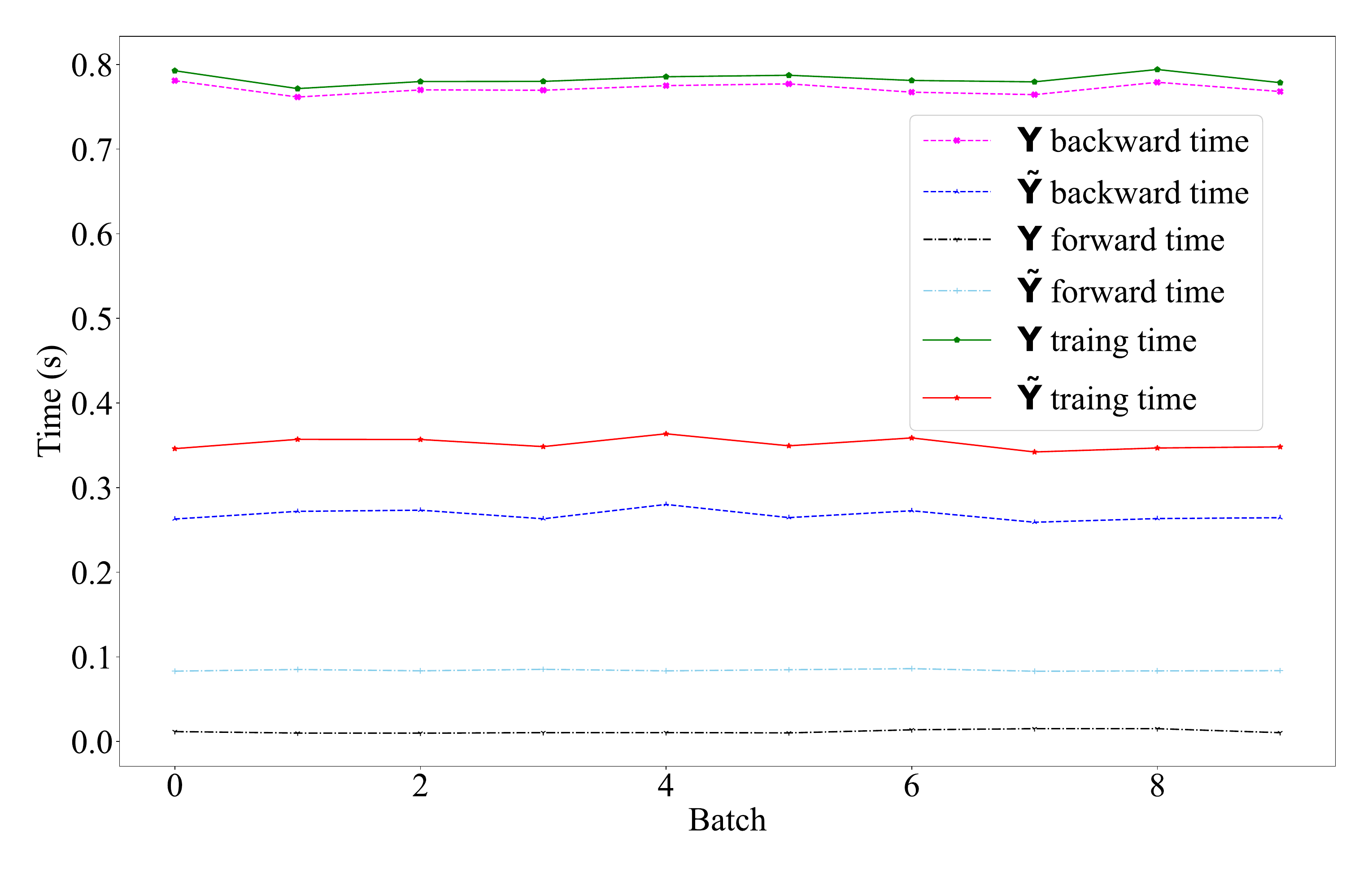}
    \caption{Training time of $\mathbf{\tilde{Y}}$ and $\mathbf{Y}$. $\mathbf{\tilde{Y}}$ corresponds to graph convolution training acceleration}
    \label{fig:speed}
    \vspace{-1em}
\end{figure}
\section{Experiments}
This section evaluates the proposed TSGCNeXt on two mainstream datasets, NTU RGB+D 60~\cite{shahroudy2016ntu} and NTU RGB+D 120~\cite{liu2019ntu}. Firstly, ablation studies are performed to prove the validity of each component. We then compare TSGCNeXt with other methods to argue for the advancement. All experiments are conducted on GeForce RTX 2080 Ti.

\subsection{Experimental Settings}
\label{sec:exp-setting}
\subsubsection{Datasets}
\noindent\textbf{NTU RGB+D 60.} NTU RGB+D60~\cite{shahroudy2016ntu} contains 56880 action videos collected by three Kinect v2 cameras with 60 indoor classes and 40 volunteers. As the last 10 classes involved two-subject interaction, $M$ is set to 2 in the experiments. The authors of this dataset recommend two benchmarks: 1) cross-subject (X-sub) contains 40320 training videos comes from 20 subjects and 16560 evaluation videos from the other 20 subjects. 2) cross-view (X-view) collects 37920 training videos from cameras 2 and 3 and 18960 evaluation videos from camera 1. Following \cite{liu2019ntu}, 302 wrong samples should be ignored. In addition to the 3D skeletal sequence of 25 joints generated by Kinect, we introduce the 2D skeletal sequence of 17 joints generated by HRNet~\cite{wang2020deep} in ST-GTCN++~\cite{duan2021revisiting} (NTU 60 hrnet2d).

\noindent\textbf{NTU RGB+D 120.} NTU RGB+D 120~\cite{liu2019ntu} is an extended version of the NTU RGB+D 60 by introducing new 60 classes, containing 114480 videos performed by 106 volunteers from three cameras views with 32 setups. There are also two benchmarks: 1) cross-subject (X-sub120) consists of 63026 training videos and 50922 evaluation videos by expanding the data on NTU RGB+D60 X-sub. 2) cross-setup (X-set120) contains 54471 videos for training with even setup IDs and 59477 videos for evaluation with odd setup IDs. According to \cite{liu2019ntu}, we ignore 532 bad samples.

\noindent\textbf{NW-UCLA.} NW-UCLA~\cite{wang2014cross} is a small-scale data set commonly used in this field to test the robustness of methods, including 1494 video clips with 10 action categories performed by 10 different subjects. It contains three views captured by Kinect cameras. Following CTR-GCN~\cite{chen2021channel}, we take data from the first two cameras as training data.
\subsubsection{Training Strategies}
We follow training technique of ConvNeXt, using AdamW ~\cite{Loshchilov2019} with a learning rate of 4e-3. The total number of training epochs is 300, with the first 20 using a linear warming technique and the remaining 280 utilizing a cosine decaying schedule. Consistent with ConvNeXt, Mixup~\cite{Zhang2018a}, Cutmix~\cite{Yun2019}, RandAugment~\cite{Cubuk2020}, and Random Erasing~\cite{Zhong2020} are used for data augmentations. The network is regularized by Stochastic Depth~\cite{Huang2016deep} and Label Smoothing~\cite{Szegedy2016a}. Besides, Exponential Moving Average (EMA)~\cite{Polyak1992} is used to improve the accuracy of the model in small batches. The 128 batches are divided equally into 2 GPUs. 
\subsection{Ablation Study}
To demonstrate the effectiveness of each module, we perform ablation studies on the bone data of NTU120 Xsub.

\begin{table}[htp]
  \centering
  \setlength{\abovecaptionskip}{0pt}
  \caption{Ablation study of graph learning mechanism.}
  \setlength{\belowcaptionskip}{0pt}
    \begin{tabular}{cccccc}
    \toprule
    Backbone & $T$ & BC & GLM &EMA Accuracy\\
    \midrule
    STGCN    & 100 &64     & classical & 86.34\\
    STGCN    & 256 &64     & classical & 85.31\\
    TSGCNeXt &100  &64     & DS-SMG & 87.06\\
    TSGCNeXt &256  &64     & DS-SMG & 88.16\\
    \midrule
    TSGCNeXt &256  &96     & none   & 79.13\\
    TSGCNeXt &256  &96     & classical & 87.11\\
    TSGCNeXt &256  &96     & SMG    & 88.41\\
    \midrule
    TSGCNeXt &100  &96     & DS-SMG & 87.63\\
    \textbf{TSGCNeXt} &\textbf{256}  &\textbf{96 }    & \textbf{DS-SMG} &  \textbf{88.86}\\
    TSGCNeXt &360  &96     & DS-SMG & 88.90\\
    \bottomrule
    \end{tabular}%
  \label{tab:Ablation Study2}%
  \vspace{-1em}
\end{table}%
\begin{table}[htp]
  \centering
  \caption{Ablation study of model structure.}
    \begin{tabular}{cccc}
    \toprule
    $k$ & SCR &EMA Accuracy\\
    \midrule
    7         & $2:5:2$ & 88.41\\
    3         & $2:5:2$ & 88.86\\
    3         & $4:3:2$ & 89.07\\
    \bottomrule
    \end{tabular}%
  \label{tab:Ablation Studyl}%
  \vspace{-1em}
\end{table}%

\begin{table*}[htbp]
  \centering
  \caption{Details of multi-stream result with $k=2$. The accuracy marked in red corresponds to the models for 2-stream fusion and the bold one is for 4-stream fusion. The batchsize for all results is 128.}
    \begin{tabular}{l|c|cc|cc}
    \toprule
    \multirow{2}[0]{*}{Benchmark} & \multirow{2}[0]{*}{Feature form} & \multicolumn{2}{c}{SCR:$2:5:2$}& \multicolumn{2}{c}{SCR:$4:3:2$}\\
              &               &    Accuracy & EMA Accuracy& Accuracy & EMA Accuracy\\
    \midrule
    \multirow{2}[2]{*}{NTU60 Xview} &  Joint  & 95.86  & \textbf{96.04} &  96.12 &  \textbf{\color{red}96.18}\\
                                    &  Bone   & 95.44  &\textbf{ 95.64} &  95.96  &   \textbf{\color{red}96.11}\\
    \midrule
    \multirow{2}[2]{*}{NTU60 Xsub} & Joint  & 91.38  & \textbf{91.64} & 91.66  & \textbf{\color{red}92.02}\\
                                   & Bone   & 91.42  & \textbf{91.91} & 91.27  & \textbf{\color{red}92.02}\\
    \midrule
    \multirow{2}[2]{*}{NTU60 hrnet2d Xview } &  Joint  & \textbf{96.82}  & 96.88 & 97.25  &   \textbf{\color{red}97.28}\\
                                             &   Bone   & \textbf{96.73}  & 96.88 & 97.01  &  \textbf{\color{red}97.21}\\
    \midrule
    \multirow{2}[2]{*}{NTU60 hrnet2d Xsub} & Joint   &  \textbf{\color{red}93.56}  &{93.51} &  93.22 &  \textbf{93.41}\\
                                           & Bone & 93.34  &  \textbf{93.44} &  93.55 & \textbf{\color{red}93.65}\\
    \midrule
    \multirow{2}[1]{*}{NTU120 Xset120} & Joint   & 89.67  & \textbf{\color{red}89.83}  &  89.29 & 89.48 \\
                                       & Bone   & \textbf{89.90} & \textbf{\color{red}90.32}  &  89.65 &  \textbf{90.25}\\
    \midrule
    \multirow{2}[2]{*}{NTU120 Xsub120} & Joint   & 87.82  & \textbf{\color{red}88.12} &  87.55 & \textbf{87.83}\\
                                       & Bone    & 88.11  &  \textbf{88.86} &  88.35 & \textbf{\color{red}89.07}\\
    \bottomrule
    \end{tabular}%
  \label{tab:bs}%
\end{table*}%

\begin{table*}
  \centering
  \caption{Results on NTU60 and 120. The comparison methods include: ST-GCN \cite{yan2018spatial}, AS-GCN \cite{li2019actional}, 2s-AGCN \cite{shi2019two}, DGNN \cite{shi2019skeleton}, Shift-GCN \cite{cheng2020skeleton}, MS-G3D \cite{liu2020disentangling}, DC-GCN+ADG \cite{cheng2020decoupling}, PA-ResGCN-B19 \cite{song2020stronger}, Dynamic-GCN \cite{ye2020dynamic}, EfficientGCN \cite{song2022constructing}, CTR-GCN \cite{chen2021channel}, ST-GCN++ \cite{duan2021revisiting}. Red text is the best result of benchmark. We have not compared the latest open source methods that have not been published. See the appendix for relevant analysis. For now, TSGCNeXt has achieved SOTA results on NTU60 X-sub (hrnet2d), X-sub120 and X-set120.}
  \label{tab:allresult}
    \begin{tabular}{lc|cc|cc|cc}
    \toprule
    \multirow{2}[0]{*}{Model} & \multirow{2}[0]{*}{Conference} & \multicolumn{2}{c}{NTU 60} & \multicolumn{2}{c}{NTU 60 hrnet2d} & \multicolumn{2}{c}{NTU 120} \\
          &       & X-sub & X-view & X-sub & X-view & X-sub120 & X-set120 \\
    \midrule
    ST-GCN & AAAI18 & 81.5  & 88.3  & $-$ & $-$ & 70.7  & 73.2 \\
    AS-GCN & CVPR19 & 86.8  & 94.2  & $-$ & $-$ & 77.9  & 78.5 \\
    2s-AGCN & CVPR19 & 88.5  & 95.1  & $-$ & $-$ & 82.9  & 84.9 \\
    DGNN (1s)  & CVPR19 & 89.2  & 95.5  & $-$ & $-$ & $-$ & $-$ \\
    DGNN  & CVPR19 & 89.9  & 96.1  & $-$ & $-$ & $-$ & $-$ \\
    Shift-GCN (1s) & CVPR20 & 87.8  & 95.1  & $-$ & $-$ & 80.9  & 83.2  \\
    Shift-GCN & CVPR20 & 90.7  & 96.5  & $-$ & $-$ & 85.9  & 87.6 \\
    MS-G3D (1s) & CVPR20 & 90.1  & 95.3  & $-$ & $-$ & $-$ & $-$ \\
    MS-G3D & CVPR20 & 91.5  & 96.2  & $-$ & $-$ & 86.9  & 88.4 \\
    DC-GCN+ADG & ECCV20 & 90.8  & 96.6  & $-$ & $-$ & 86.5  & 88.1 \\
    PA-ResGCN-B19 & ACMMM20 & 90.9  & 96.0   & $-$ & $-$ & 87.3  & 88.3 \\
    Dynamic-GCN & ACMMM20 & 91.5  & 96.0   & $-$ & $-$ & 87.3  & 88.6 \\
    EfficientGCN & TPAMI22 & 92.1  & 96.1  & $-$ & $-$ & 88.7  & 88.9 \\
    CTR-GCN (1s) & ICCV2021 & 90.6  & $-$ & $-$ & $-$ & 85.7  & 87.5  \\
    CTR-GCN (2s) & ICCV2021 & $-$ & $-$ & $-$ & $-$ & 88.7  & 90.1 \\
    CTR-GCN & ICCV2021 & 92.4  & 96.8  & $-$ & $-$ & 88.9  & 90.6 \\
    ST-GCN++ (1s) & CVPR2022 & 90.1  & 95.6  & 92.3  & 97.4  & 85.6  & 87.5  \\
    ST-GCN++ (2s) & CVPR2022 & 91.4  & 96.7  & 92.8  & 98.4  & 87.0  & 89.1 \\
    ST-GCN++ (4s) & CVPR2022 & 92.1  & 97.0  & 93.2  & \color{red}98.5  & 87.5  & 89.8 \\
    \midrule
    TSGCNeXt (1s) &  $-$    & 92.02 & 96.18 & 93.65 & 97.28 & 89.07 & 90.32\\
    TSGCNeXt (2s) & $-$     & 92.91 & 96.87 & 94.27 & 97.76 & 89.98 & 91.46\\
    TSGCNeXt (new4s) & $-$  & 93.16 & 96.98 & \color{red}94.47 & 97.82  & \color{red}90.22 & \color{red}91.74\\
    \bottomrule
    \end{tabular}%
  \label{tab:finnal}%
\end{table*}

\noindent\textbf{Graph learning mechanism.} The section verifies long-term learning ability and effectiveness of DS-SMG. We train each verification model under the same training settings for fairness. STGCN with frame $T=100$ is selected as the baseline. To compare the effectiveness of the long time series, we increase $T$ to 256 and train baseline. To fairly compare the DS-SMG under the same channel setting, we train TSGCNeXt under the condition of base channels(BC) equals $64$ with $T=100$ and $T=256$. Further, we increase $T$ to 360 to verify the long-term learning potential. For the effectiveness of DS-SMG comes from the overall module, we first remove the graph learning mechanism and only use the module of ConvNeXt for training. Then we use the classical graph learning mechanism to replace the DS-SMG module for training. Further, we remove static branches to prove the effectiveness of DS. 

 In Tab.~\ref{tab:Ablation Study2}, the accuracy of STGCN is reduced with the frame increasing. TSGCNeXt is superior to STGCN in each time series configuration and still has space to rise accuracy for longer time series. This means that TSGCNeXt has greater potential for processing long-term data. TSGCNeXt reaches 88.90\% accuracy on $T=360$, but we finally choose $T=256$ as the primary setting to match previous works as shown in Fig.\ref{fig:longtime}. For the GLM, the design of DS-SMG promotes accuracy. Relying solely on the structure of ConvNeXt is not an effective way to learn skeleton graph features without any learning mechanisms. SMG alone also has a 1.3\% improvement over classical GLM. With the introduction of DS, the accuracy is further improved 0.45\%.

\noindent\textbf{Temporal receptive field.} The benchmark model references ConvNeXts, $k=7$ and SCR $2:5:2$. We reduce $k$ to 3. The accuracy is improved to 88.86\% with SCR $2:5:2$ and 89.07\% with SCR $4:3:2$.

\noindent\textbf{Training speed.} To prove the effectiveness of the graph convolution training acceleration, we count the time of forward and backward calculation in a epoch in batch. As shown in the Fig.4, the backward of $\mathbf{\tilde{Y}}$ takes up almost 66\% 
less time than $\mathbf{Y}$. Although the forward time of $\mathbf{\tilde{Y}}$ is longer than that of $\mathbf{Y}$, from the perspective of the whole training time, $\mathbf{\tilde{Y}}$ is still faster than $\mathbf{Y}$ with 55.08\% speed-up.

\subsection{Experimental Results}
\label{er}
\noindent\textbf{Results and Comparison.} To fully demonstrate the performance of TSGCNeXt, we train multiple streams (${{\mathcal{X}}_{\text{j}}}$, ${{\mathcal{X}}_{\text{b}}}$, ${{\mathcal{X}}_{\text{jm}}}$, ${{\mathcal{X}}_{\text{bm}}}$) on 3D skeletal data (NTU 60), HRNet2d skeletal data (NTU 60 hrnet2d) from NTU 60 and 3D skeletal data from NTU 120 (NTU 120). We compare TSGCNeXt with SOTA methods to show the advantage. 

First, the accuracy of all model setting is reported in the Tab.~\ref{tab:bs}.  EMA will bring a certain gain to the recognition accuracy. We mark the stage compute rates corresponding to high EMA accuracy in each data joint and bone stream in red using for two stream fusion. Different features have different preferences for stage compute ratios. In addition, the performance of bone stream and joint stream is also inconsistent on multiple datasets. Therefore, the aggregation of multi stream information can unify these differences and effectively improve the accuracy of recognition.

In the previous work, the fusion of multiple streams usually adopts joint stream, bone stream, joint motion stream and bone motion stream. But, we introduce the convergence of networks with different structures, as shown in Tab.~\ref{tab:bs}. In addition to NTU120 Xset120 replacing the Joint EMA (stage compute ratios $=4:3:2$) with a more accurate Bone stream (stage compute ratios $=2:5:2$) , we aggregate the models with higher single stream accuracy in the Joint and Bone designed with two model structures.

With the new four stream fusion strategy, we have achieved higher accuracy, as shown in Tab.~\ref{tab:allresult}. The results include the accuracy of two streams (2s) and four streams (4s). The accuracy of motion features is generally lower than that of non-motion, so this flow is not used in four flow models. The results are shown in Tab.~\ref{tab:finnal}. TSGCNeXt can achieve an accuracy comparable to the current SOTA at two streams. On the NTU60, the 2s accuracy is higher than that of the best ST-GCN++. On NTU120, the two-stream accuracy exceeds the 4-stream accuracy of the best CTR-GCN. However, the NTU60 X-view accuracy is slightly lower than that of ST-GCN++ at 4-stream. This is because the accuracy metric is less challenging to learn, and the accuracy tends to be saturated. Our method has improved significantly on other more complex metrics, especially on the most difficult-to-learn NTU120. In the three rankings of NTU60 X-sub, X-set120 and X-sub120, our new four stream integration results have outstanding performance, reaching 94.47\%, 91.74\% and 90.22\%, respectively.

In addition, to further validate the effectiveness of TSGCN, we conduct comparison experiments on small-scale data NW-UCLA. As shown in Tab.~\ref{tab:NW}, the results are also better than those already published before.
\begin{table}[htp]
  \centering
  \setlength{\abovecaptionskip}{0pt}
  \caption{Result on NW-UCLA.}
  \setlength{\belowcaptionskip}{0pt}
    \begin{tabular}{cccccc}
    \toprule
    Model &Accuracy\\
    \midrule
    AGC-LSTM~\cite{si2019attention}    & 93.3\\
    Shift-STGCN~\cite{cheng2020skeleton}     & 94.6\\
    DC-GCN+ADG~\cite{cheng2020decoupling}   & 95.3\\
    CTR-GCN~\cite{chen2021channel}     & 96.5\\
    \midrule
    TSGCNeXt  & 96.55\\
    \bottomrule
    \end{tabular}%
  \label{tab:NW}%
  \vspace{-1em}
\end{table}%

\noindent\textbf{Discussion.} Efficient training and long-term learning capability are the most important features of TSGCNeXt. Despite the larger channel setting and longer time series in the design, as shown in Fig.~\ref{fig:longtime}, our method has fewer parameters and faster training speed. The accuracy of TSGCNeXt is still improving on longer time series without considering the limitation of the parameters. This effectively addresses the bottleneck of long-time training and demonstrates the long-term learning potential. Further, the simple yet effective structural design opens up new horizons for subsequent work, rather than stacking complex learning mechanisms.

\section{Conclusion}
We propose the Temporal-Spatio Graph ConvNeXt, effectively addressing the parameter and efficiency bottleneck in long time series skeletal action recognition. A simple and effective graph learning mechanism, DS-SMG, is proposed to avoid the loss of node information during dynamic graph learning. In addition, TSGCNeXt can learn efficiently for long time sequences with the support of a graph convolutional training acceleration strategy and a long time-series friendly overall network design. Through plenty of experiments, TSGCNeXt is superior to the SOTA method. In future work, we will optimize the TSGCNeXt parameter design and network structure, exploring the application of DS-SMG in other graph-related tasks.
{\small
\bibliographystyle{ieee_fullname}
\bibliography{egbib}
}

\newpage

\appendix

\section{Appendix}
\subsection{Hyper-parameters Selection}
\label{app:hyper-parameters-select}
In this subsection, we describe the specific parameter settings in the experiment. As shown in the Tab.~\ref{tab:train_detail}, the generic parameters are basically the same as ConvNeXt. 
\begin{table}[htbp]
  \centering
  \caption{A generic hyperparameter on each model.}
    \begin{tabular}{l|cc}
    \toprule
    training config & ImageNet-1K \\
    \midrule
    weight init & trunc. normal (0.2) \\
    optimizer & AdamW\\
    base learning rate & 4e-3 \\
    weight decay & 0.05  \\
    optimizer momentum & $\beta_1, \beta_2{=}0.9, 0.999$\\
    training epochs & 300\\
    learning rate schedule & cosine decay\\
    warmup epochs & 20\\
    warmup schedule & linear\\
    layer-wise lr decay & None \\
    randaugment & (9, 0.5)\\
    mixup & 0.8\\
    cutmix & 1.0 \\
    random erasing & 0.25\\
    label smoothing & 0.1\\
    layer scale & 1e-6\\
    exp. mov. avg. (EMA) & 0.9999\\
    drop\_out & 0.4\\
    \bottomrule
    \end{tabular}%
  \label{tab:train_detail}%
  \vspace{-1em}
\end{table}%

\subsection{Experimental details}
\label{app:Multi-stream result setting}
In this section, we mainly supplement the information in Tab.~\ref{tab:bs} of the body. Details mainly focus on the experimental results of two stage compute ratios with $batchsize=128$. We select four models as shown in the Tab.~\ref{tab:ws} for fusion with weight. 

To show advantage, we compare some current SOTA work (including some work that has been open source but has not been published) as shown in Tab.~\ref{tab:result}. The results show that TSGCNeXt has the best results in three of the four lists. Although it is lower than ST-GCN (hrnet2d) on the NTU-60 X-view list, our other three indicators are better. This fully demonstrates the superiority of our method.

\begin{table}[htbp]
  \centering
  \caption{The weight ration for each stream. The streams marked in red are enabled for 2 streams.}
    \begin{tabular}{l|cc|c}
    \toprule
    Benchmark & SCR &Feature form & weight \\
    \midrule
    \multirow{4}[2]{*}{NTU60 Xview}&$2:5:2$ & Joint EMA & 0.70  \\
          & $2:5:2$& Bone EMA & 0.40  \\
          &\color{red}{$4:3:2$}  &\color{red}{ Joint EMA}& 1  \\
          &\color{red}{$4:3:2$}  & \color{red}{Bone EMA}& 0.8  \\
    \midrule
    \multirow{4}[2]{*}{NTU60 Xsub}& $2:5:2$& Joint EMA & 0.05  \\
         & $2:5:2$& Bone EMA  & 0.85  \\
         &\color{red}{$4:3:2$ } & \color{red}{Joint EMA} & 1.00  \\
         &\color{red}{$4:3:2$}  &\color{red}{ Bone EMA }& 0.75  \\
    \midrule
    \multirow{4}[2]{*}{NTU60 hrnet2d Xview }& $2:5:2$& Joint EMA & 0.50  \\
         & $2:5:2$&Bone EMA& 0.30  \\
         &\color{red}{$4:3:2$  }& \color{red}{Joint EMA} & 1.00  \\
         &\color{red}{$4:3:2$}  & \color{red}{Bone EMA} & 0.95  \\
    \midrule
    \multirow{4}[2]{*}{NTU60 hrnet2d Xsub}& {$2:5:2$}& Joint & 0.95  \\
         & $2:5:2$& Bone EMA  & 0.90  \\
         &\color{red}{$4:3:2$}  & \color{red}{Joint EMA}& 0.90  \\
         &\color{red}{$4:3:2$}  & \color{red}{Bone EMA} & 1.00  \\
    \midrule
    \multirow{4}[2]{*}{NTU120 Xset120}&\color{red}{$2:5:2$} & \color{red}{Joint EMA} & 0.95  \\
        & \color{red}{ $2:5:2$} & \color{red}{Bone EMA}  & 1.00  \\
        &$2:5:2$   & Bone & 0.85  \\
        &$4:3:2$   & Bone EMA & 0.30  \\
    \midrule
    \multirow{4}[2]{*}{NTU120 Xsub120}&\color{red}{$2:5:2$} & \color{red}{Joint EMA} & 0.90  \\
        & $2:5:2$ & Bone EMA  & 0.70  \\
        &$4:3:2$   & Joint EMA & 0.55  \\
        &\color{red}{$4:3:2$   }& \color{red}{Bone EMA} & 1.00  \\
    \bottomrule
    \end{tabular}%
  \label{tab:ws}%
\end{table}%

\begin{table}
  \centering
  \caption{Results compare with SOTA.}
  \label{tab:result}
    \begin{tabular}{l|cc|cc}
    \toprule
    Model & \multicolumn{2}{c}{NTU 60} & \multicolumn{2}{c}{NTU 120} \\
         & X-view &X-sub & X-set120 & X-sub120 \\
    \midrule	
    STGAT &97.3& 92.8& 90.4 & 88.7\\
    ST-GCN++ &98.3&92.4&89.0&84.7\\
    PoseC3D&97.1&94.1&-&-\\
    PSUMNet &96.7 &92.9& 90.6 & 89.4\\
    TCA-GCN &97.0 &92.8 & 90.8 & 89.4\\
    LST &97.0 &92.9 & 91.1 & 89.9\\
    HD-GCN &97.2 &93.4 & 91.6 & 90.1\\
    \midrule
    TSGCNeXt& 97.82 & \color{red}94.47 & \color{red}91.74 & \color{red}90.22 \\
    \bottomrule
    \end{tabular}%
  \label{tab:newcompare}%
\end{table}

\end{document}